\documentclass[twoside,11pt]{article}

\usepackage{framed,multirow}	
\usepackage{jmlr2e}
\usepackage{amsmath} 
\usepackage{breqn}
\usepackage{algorithm}
\usepackage{algorithmic}
\usepackage{amsfonts}
\usepackage{booktabs}
\usepackage{siunitx}
\usepackage{graphicx}
\usepackage{threeparttable}
\usepackage{rotating, graphicx}
\usepackage{multirow} 
\usepackage{xcolor}

\usepackage{xcolor}

\definecolor{brilliantrose}{rgb}{1.0, 0.33, 0.64}
\definecolor{darkpink}{rgb}{0.91, 0.33, 0.5}
\newcommand{\MYhref}[3][darkpink]{\href{#2}{\color{#1}{#3}}}

\jmlrheading{22}{2021}{1-25}{4/21}{00/00}{meila00a}{Caroline Pacheco do Espírito Silva, José A. M. Felippe De Souza, Antoine Vacavant, Thierry Bouwmans, Andrews Cordolino Sobral}

\ShortHeadings{Automated Mathematical Equation Structure Discovery for Visual Analysis}{}
\firstpageno{1}

\begin{document}

\title{Automated Mathematical Equation Discovery for Visual Analysis}

\author{\name Caroline Pacheco do Espírito Silva \email lolyne.pacheco@gmail.com \\
       \addr Machine Learning Team, ActiveEon\\
       \addr Paris, France
       \AND
       \name José A. M. Felippe De Souza \email felippe@ubi.pt\\
       \addr Department of Electromechanical Engineering\\
       University of Beira Interior\\
       Covilha, Portugal
       \AND
       \name  Antoine Vacavant \email antoine.vacavant@uca.fr \\
       \addr Institut Pascal, Université Clermont Auvergne\\
       UMR6602 UCA/SIGMA/CNRS, F-63171 \\
       Aubière, France
       \AND
       \name Thierry Bouwmans \email thierry.bouwmans@univ-lr.fr \\
       \addr Laboratoire de Mathématiques, Image et Applications\\
       Université de La Rochelle\\
       La Rochelle, France
       \AND
       \name Andrews Cordolino Sobral \email andrewssobral@gmail.com\\
       \addr Machine Learning Team, ActiveEon\\
       \addr Paris, France}

\editor{}

\maketitle

\begin{abstract}
Finding the best mathematical equation to deal with the different challenges found in complex scenarios requires a thorough understanding of the scenario and a trial and error process carried out by experts. In recent years, most state-of-the-art equation discovery methods have been widely applied in modeling and identification systems. However, equation discovery approaches can be very useful in computer vision, particularly in the field of feature extraction. In this paper, we focus on recent AI advances to present a novel framework for automatically discovering equations from scratch with little human intervention to deal with the different challenges encountered in real-world scenarios. In addition, our proposal can reduce human bias by proposing a search space design through generative network instead of hand-designed. As a proof of concept, the equations discovered by our framework are used to distinguish moving objects from the background in video sequences. Experimental results show the potential of the proposed approach and its effectiveness in discovering the best equation in video sequences. The code and data are available at: \emph{\textbf{\MYhref{https://github.com/carolinepacheco/equation_discovery_scene_analysis}{https://github.com/carolinepacheco/equation\_discovery\_scene\_analysis}}}. 
\end{abstract}


\section{Introduction}

In the past few decades, the availability of big data, computing power and the breakthroughs in machine learning algorithms have dramatically increased the impact of artificial intelligence (AI) on our society. AI has been successfully applied to solve many challenging tasks, in areas such as computer vision~\citep{deng:2009, marck:2010, krizhevsky:2012, shi:2016, hsu:2018, shaham:2019, zhengqi:2019, radosavovic2019, xie:2019}, speech recognition~\citep{dahl:2013, maas:2014, bukhari:2017, hadian:2018, stephenson:2019, schneider2019}, natural language processing~\citep{sutskever:2014, bahdanau:2015, wu:2016, changmeng:2019, lewis2020}, and others. Despite these impressive advances, significant human expertise is still needed to design promising AI algorithms to solve complex task in the most diverse areas.

Until a few years ago, AI algorithms were able to successfully learn to recognize places, objects, or people but they were not good at creating their own plausible data. Nevertheless, everything started to change with the advances in Deep Generative Networks. They try to simulate the ``innovation potential" of the human brain by learning the true data distribution of the training set, in order to generate realistic synthetic data in the form of images,  video, audio or text.  Variational Autoencoders (VAE)~\citep{kingma:2014}, Generative Adversarial Networks (GAN)~\citep{goodfellow:2014} and their variants are the most popular types of generative models. The basic idea of the VAE is to maximize the lower bound of the data log-likelihood while GAN aims to achieve a balance between Generator $G$ that generates real-like samples from a random noise input and Discriminator $D$ that tries to differentiate $G^'s$ generated samples from real training data.

Another important breakthrough in the field of machine learning has been the Automated Machine Learning (AutoML). It is not a new concept, the AutoML techniques have been explored by the scientific community since the 1990s~\citep{ripley:1993, king:1995, kohavi:1995, michie:1995,  guyon:2003, chen:2004,  samanta:2004, guyon:2008, escalante:2009, bergstra:2012, feurer:2015, hoffer:2017}. Nevertheless, the AutoML techniques have recently received considerable attention from both scientific and industrial communities. This can be explained by the the exponential growth in computing power and data storage capacity that made it possible to automate machine learning pipelines - from pre-processing, feature extraction, feature selection, model fitting and selection, and hyper-parameter tuning with minimal human intervention ~\citep{drori:2018, hutter:2019, feurer:2019}. AutoML's popularity has increased further still since 2016 when Google Brain team revealed their first Neural Architecture Search (NAS)~\citep{zoph:2017}. It a subfield of AutoML that allows to automate the manual process to discover the optimal network architecture, which significantly reduces human labor. Given a human-designed search space containing a set of possible neural networks architectures, NAS uses an optimization method to automatically find the best combinations within the search space~\citep{kobler:2009, zoph:2017, liu:2018, hutter:2019}. Despite the effectiveness of NAS, it still remains laborious to define a useful search space, because it often requires human expert knowledge and involves a process of trial-and-error. In addition, the search space usually relies on human bias, as its design depends on human-designed architectures as starting points.

Currently, research in the field of NAS focuses mainly on discovery of high-performance neural network architectures that were predefined by humans. However, reducing the search space of the human bias can help increase of the innovation potential of NAS, meaning a great breakthrough in the development of AI systems that can create novelties with high performance on their own without human intervention~\citep{hutter:2019, real:2020}. An impressive research recently presented by~\citep{real:2020}, showed that AutoML can discover completely new algorithms through a generic search space. They proposed a framework for discovering machine learning algorithms using regularized evolution search method. Inspired by their research, in this paper we propose a novel framework for discovering  mathematical equations that even human developers may not have thought of yet with minimal human intervention. Given a search space designed by VAE instead hand-designed containing a set of possible Local Binary Patterns (LBP) equation structures, the Multi-Objective Covariance Matrix Adaptation Evolution Strategy (MO-CMA-ES)~\citep{igel:2007, hansen:2011} searches the best LBP equation by mutating its arithmetic operators. Next, the performance of each LBP equation is estimated in a background subtraction problem ~\citep{sobral:2014} to distinguish the moving objects from a set of videos. Finally, the LBP equation which presented the maximum precision is selected as a best equation for dealing with specific challenges (e.g., changes in lighting, dynamic backgrounds, noise, shadows and others) in a real-world scene.

In summary, our contributions are as follows:

\begin{itemize}
	\item a search space designed by VAE containing a wide variety of mathematical equation structures that we may never have thought of yet;
	\item a new framework capable of discovering the most suitable LBP mathematical equations to deal real-world scenarios;
	\item an extensive experiments to prove the potential of our approach to distinguish moving objects from the background in video sequences.
\end{itemize}

The remainder of the paper is organized as follows. \hyperref[sec:related_work] {Section II} provides quite a comprehensive overview of state-of-the-art of the equation discovery approaches. \hyperref[sec:propoded_method]{Section III} introduces our framework by illustrating it on a background subtraction problem.
\hyperref[sec:experimental_reults]{Section IV} shows the experimental results which are reported and discussed. Finally, \hyperref[sec:conclusion]{Section V} concludes the paper with a summary and a brief outline of some directions for further research.

\begin{figure}[t]
\label{fig:work_graph_papers}
 \centering
  \includegraphics[width=0.8\textwidth]{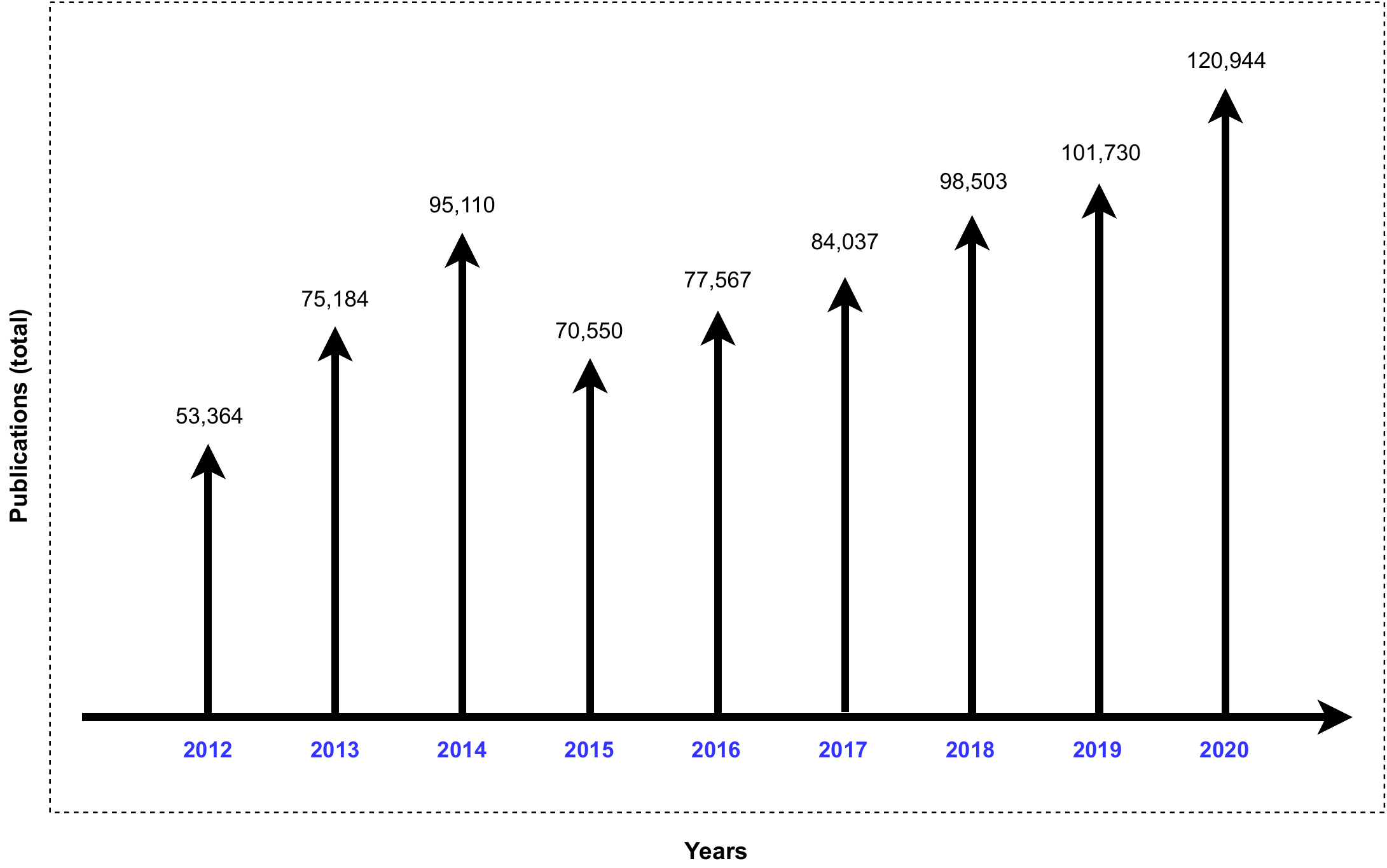}
 \caption{\small Scientific publications based on LBP from 2012, the year the Convolutional Neural Networks (CNN) achieved a top-5 error of 15.3 in the ImageNet 2012 Challenge, up to 2020. In total, 776.989 LBP research papers were proposed in this period 2012-19 showing the growing interest of the scientific community in the subject. Source: \cite{dimensions:2021}.}
\centering
\end{figure}

\section{Related Work}
\label{sec:related_work}

Equation discovery is a topic in the machine learning area that automates the process of formulating mathematical equations to find the equation that best fits through data ~\citep{langley:1995, mitchell:1997, todorovski:2001}. It has achieved remarkable success late 1980s, since publication of the BACON system proposed by~\cite{langley:1987, langleyp:1987}. Six versions of BACON have been proposed to rediscover various physical laws, such as the gravitation's law, the Boyle’s law, the Ohm’s law and others - from experimental data. All BACON system versions share a common approach, as well as a common representation of data and laws for equation discovery. The differences between them are in the set of data-driven heuristics that is used to search for empirical laws. 
One of the main limitations of the BACON systems is that they do not support irrelevant variables in the input data. Another pioneering system in discovery equations that deals with this restriction is the ABACUS~\citep{falkenhainer:1986}. In contrast to BACON, ABACUS allows irrelevant variables to be present in the input data, making the combination of existing variables to formulate new intrinsically exponential variables. To overcome this limitation, ABACUS prohibits mathematically redundant expressions and physically impossible relationships. Other related methods that extend or use approaches similar to BACON and ABACUS are FAHRENHEIT~\citep{zytkow:1987}, IDS ~\citep{nordhausen:1990} ~\citep{shrager:1990} and KEPLER ~\citep{wu:11989}, Coper ~\citep{kokar:1986, kokarm:1986} and Fortyniner~\citep{zytkow:1991}.

~\cite{dzeroski:1995} proposed LAGRANGE, the first equation discovery approach to dealing with differential equations. It exhaustively searches for the space of polynomial equations with limited number of terms that are based on observed system variables and their time derivatives. LAGRANGE allows the user to define the space of candidate equations by selecting the parameter values, such as the maximum number of terms or maximum degree of the polynomial. Two years later,~\cite{todorovski:1997} introduced the successor to the LAGRANGE framework, the LAGRAMGE. Instead of using the model structure provided by a human expert, like most other methods, LAGRAMGE relies on context free grammars to specify a wide range of possible equation structures that make sense from the expert's point of view. It allows expressing different types of knowledge by grammar, however, it is not very easy for a domain expert to encode their domain knowledge in the form of grammar. Besides, the grammar-based formulation is generally task specific, so a grammar created to model a certain system cannot necessarily be used to model other systems in the same domain. Some years later,~\cite{trodorovski:2007} extend the LAGRAMGE method by proposing the LAGRAMGE 2.0 version.  It provides a formalism for integrating knowledge of population dynamics modeling for discovering equations. This formalism allows the codification of a high level domain knowledge accessible to human specialists. Coded knowledge can be automatically transformed into operational form of context-dependent grammars. The experimental results show that the integration of specialized knowledge within the equation discovery process significantly improves the robustness and efficiency of the LAGRAMGE 2.0 method when compared to the previous version. LAGRAMGE framework has been successfully applied in different real-world applications. For instance, ~\cite{ganzert:2010} adapt the LAGRAMGE method to discover mathematical equations that model a mechanically ventilated lung. In~\citep{alzaidi:2011}, the method is used to find equations that predict London Metal Exchange commodity prices. ~\cite{markic:2013} also use this method to discover equations in the field of earthquake engineering to predict peak ground acceleration.

~\cite{simidjievski:2019} recently presented a process-based approach to discover equations for modeling nonlinear dynamic systems from measurements and domain-specific modeling knowledge. Although the approach has been applied to a cascaded water tank system with two tanks, the authors claim that the proposed method can be applied in a variety of other system identification tasks.~\cite{tanevski:2020} shows a knowledge-driven strategy for selecting a suitable structure from a finite set of user-specified equation structures using various combinatorial search algorithms. After carrying out different experiments in reconstructing the structure of dynamical systems models, the authors noticed that using the combination of beam~\citep{bisiani:1992} and tabu~\citep{glover:1997} search algorithms or particle swarm optimization~\citep{parsopoulos:2010}, achieves the best trade-off between speed of convergence and recall. Nonetheless, search algorithms with moderate diversification are the most suited algorithms to biased equation discovery. Finally, the authors conclude that genetic algorithms showed a lower performance for knowledge-based equation discovery.

Most state-of-the-art equation discovery methods have been widely applied in the modeling and identification systems. However, we believe that equation discovery approaches can be very useful in the scope of computer vision, particularly in the field of feature extraction. Despite the great success of deep learning techniques ~\citep{krizhevsky:2012, lecun:2015, goodfellow:2016}, a variety of texture extraction methods have recently attracted great attention for different  applications, especially the Local Binary Pattern (LBP)~\citep{ojala:2002} because it is simple and quick to calculate. Figure 1 
shows a graph of the LBP-based papers proposed since 2012, the year the Convolutions Neural Network (CNN) achieved a top-5 error of 15.3 \% in the ImageNet 2012 Challenge, up to 2020. In total 776.989 LBP papers were proposed in the period 2012-19 showing the growing interest of the scientific community for the subject. LBP is a texture descriptor which labels the pixels of an image by establishing a threshold in the neighborhood of each pixel and considers the result as a binary number. It shows great invariance to changes in monotonic lighting, does not require many parameters to be set, and has a high discriminative power. However, sometimes the original LBP equation needs to be reformulated to make it capable to deal with several challenges (e.g., sudden changes in lighting, dynamic backgrounds, camera jitter, noise, shadows, and others) found in different types of scenes. Nevertheless, to formulate a mathematical LBP equation that is robust to the different challenges faced in a complex scene, in-depth knowledge of the scene and trial-error process by experts is required. Therefore, in this paper we focus on recent AI advances to propose a framework to automatically discovering LBP equations from scratch with little human intervention to deal with different challenges encountered in complex scenes.

\section{Proposed Method}
\label{sec:propoded_method}

The proposed approach to discovering equations consists of two main components: (1) Discover the best VAE to design our search space that contains a wide variety of mathematical LBP equation structure that we may never have thought of yet. (2) Search the best LBP equation to deal with different challenges (e.g., changes in lighting, dynamic backgrounds, camera jitter, noise, shadows, and others) faced by a background subtraction algorithm in detecting moving objects found in complex scenarios. We describe each of these components in the remainder of this section.

\subsection{Search Space Design by VAE to Generate Equation Structures}
\label{sec:search_space_design}

The search space design is a key component for equation discovery. In the search space, we must define which equation structures will be designed and optimized. Instead of hand-designed search space, we design it through a Variational Autoencoder (VAE). Given a set of parameters denoted by $S$, the MO-CMA-ES (see subsection \ref{subsec:search_strategy_best_equation} for further details) is responsible for generating a set of parameters to train different VAE's expressed by $\Psi = \{ \psi_1, \psi_2, \ldots, \psi_L \}$, where $\psi$ represents a VAE instance and $L$ is the maximum number of elements $\psi \subseteq \Psi$. Next, the VAE with the optimum parameters that allow best well-performing is selected. Finally, the $\psi \subseteq \Psi$ which presented the maximum accuracy $\{ \arg\max_{j \in 1, \ldots, L} P_{j}(x)\}$ is chosen to design our search space defined as $E$ that will contain a set of new LBP equations. The $P$ defines a set of accuracy for each $\psi \subseteq \Psi$. A brief overview of this first step of our proposed framework is illustrated in Figure 2. The VAE is introduced below.

\begin{figure}[t]
\label{fig:framework_part1}
 \includegraphics[width=\linewidth]{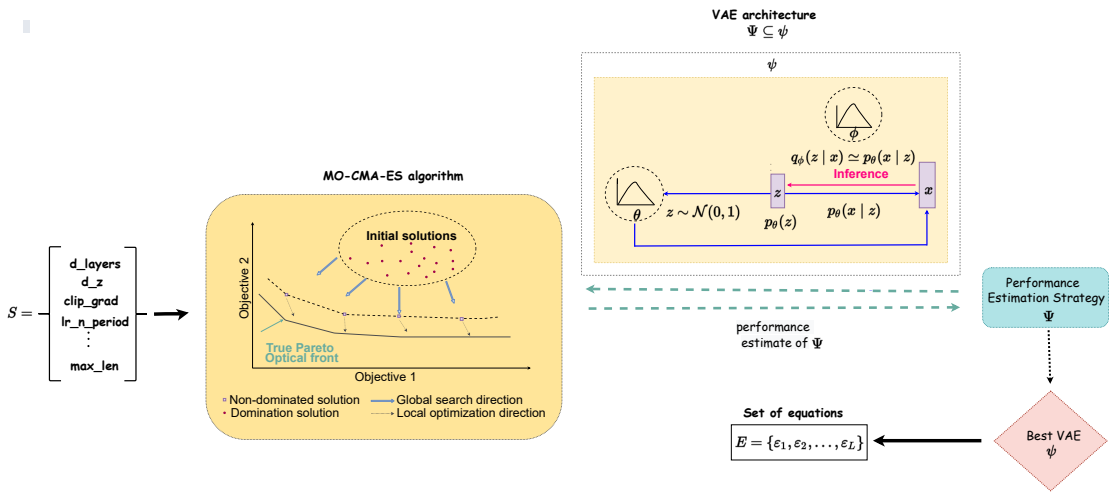}
 \caption{\small Brief overview of the first component of the proposed framework. Given a set of parameters denoted by $S$,  the MO-CMA-ES is responsible for generating a set of optimal parameters to train different VAE's expressed by $\Psi = \{ \psi_1, \psi_2, \ldots, \psi_L \}$, where $\psi$ represents a VAE and $L$ is the maximum number of elements $\psi \subseteq \Psi$. Then, the VAE with the optimum parameters that allow best performance is selected. Finally, the $\psi \subseteq \Psi$ which presented the highest performance is chosen to design the search space $E$.}
\centering
\end{figure}


\subsubsection{Variational Autoencoder (VAE)}
Variational Autoencoder (VAE) is a neural network-based generation model that learns an encoder and a decoder that leads to data reconstruction and the ability to generate new samples that share similar statistics from the input data. Given a small set of equations  $x$ the VAE maps the $x$ onto a latent space with a probabilistic encoder $q_\phi(z \mid x)$ and reconstructs samples with a probabilistic decoder $p_\theta(x \mid z)$. A Gaussian encoder was used with the  following reparameterization trick:


\begin{equation}
\label{eq:1}
q_\phi(z \mid x) =  \mathcal{N}(z \mid  \mu_{_\phi}(x), \sigma_{_\phi}(x)) = \mathcal{N}(\in \mid 0, I) \cdot \sigma_{_\phi}(x) + \mu_{_\phi}(x)
\end{equation}

Gaussian distribution $\mathcal{N}(0, I)$ is the most popular choice for a prior distribution $p{_\psi}(z)$ in the latent space. The  Kullback-Leibler divergence, or simply, the $\mathcal{KL}$ divergence is a similarity measure  commonly used to calculate difference between two probability distributions. To ensure that $q(z \mid x)$ is similar to $p(x \mid z)$, we need to minimize the $\mathcal{KL}$ divergence between the two probability distributions. 


\begin{equation}
\label{eq:2}
\min \mathcal{KL}(q_{_\phi} (z \mid x) \| p_{\theta} (z \mid x)) 
\end{equation}


We can minimize the above expression by maximizing the following:

\begin{equation}
\label{eq:3}
L(\phi, \theta: x) =  E_{z \sim q_{_\phi} (z \mid x)}(\log ( p_{\theta} (x \mid z)) - \mathcal{KL}(q_{_\phi} (z \mid x) \| p (z)) 
\end{equation}

As we see in Eq.(\ref{eq:3}), the loss function for VAE consists of two terms, the \textbf{\textit{reconstruction term}} that penalizes error between the reconstruction of input back from the latent vector and the \textbf{\textit{divergence term}} that encourages the learned distribution $q_{_\phi} (z \mid x)$ to be similar to the true prior distribution $p(z)$, which we assume following a unit Gaussian distribution, for each dimension $j$ of the latent space.

\begin{figure}[t]
\label{fig:framework1}
 \includegraphics[width=\textwidth]{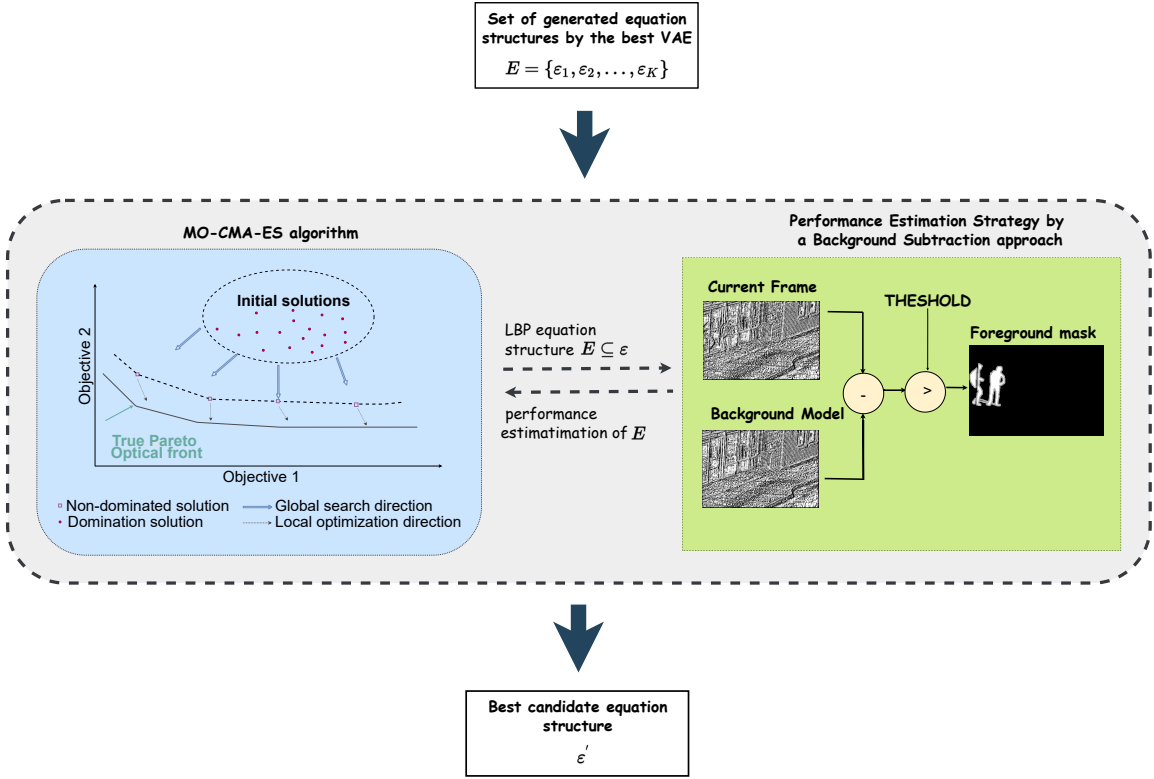}
 \caption{\small Brief overview of the second component of the proposed framework. A set of LBP equations $E = \{\varepsilon_1,  \varepsilon_2, \dots,  \varepsilon_K \}$, where $\varepsilon$ expresses each equation structures and $k$ is a user parameter that determines the number of elements $\varepsilon \subseteq E$. Initially, the MO-CMA-ES seeks the best LBP equation by mutating the arithmetic operators of each equation $\varepsilon \subseteq E$ resulting in a new one of the mutated equations $E^{'} = \{\varepsilon^{'}_1,  \varepsilon^{'}_2, \dots,  \varepsilon^{'}_K \}$. The performance of each LBP equation is estimated by a background subtraction algorithm to distinguish the moving objects from the background of a set of videos. Finally, the $\varepsilon^{'}$ that presented the maximum accuracy is selected as a best equation structure.}
\centering
\end{figure}

\begin{algorithm}[t]
	\caption{\large Equation Structure Discovery}
	\label{alg:structure_discovery}
   \normalsize
	\begin{algorithmic}[1]
    \STATE \textbf{Require:}  $S = \{{\text{ENC\_LAYERS, DEC\_LAYERS, ENC\_HIDDEN, DEC\_HIDDEN, etc.}}\}$, training set images $X$, user parameter $L$, VAE instance $\nu$, MO-CMA-ES instance, arithmetic operators  $A = \{+ ,  -, *, /\}$, user parameter $K$
	\STATE \textit{l} $\leftarrow 1$
	\STATE \textbf{repeat}
	\FOR{$l = 1:L$}
	\STATE // search the best VAE $\psi$
	\STATE  $\psi_{l} \leftarrow$ \textit{MO-CMA-ES ($\nu$ ,$S$)} 
	\STATE $P_{l} \leftarrow$ \textit{Performance-Estimation($\psi_{l})$}
    \STATE \textit{vae-scores} $\leftarrow$ \{$P_{l}$, $\psi^{'}_{l} \} $
	\ENDFOR	
	\STATE // Select the best VAE
	\STATE  \{$P$, $\psi \} $ = $\arg\max(\textit{vae-scores})$
    \STATE
    \STATE // generate a set of LBP equation structure $E= \{ \varepsilon_1, \varepsilon_2, \dots, \varepsilon_K \}$ using the best VAE $\psi$
	\STATE \textit{k} $\leftarrow 1$
	\STATE \textbf{repeat}
	\FOR{$k = 1:K$}
	\STATE // generate a set of LBP equations  $\varepsilon^{'}$ by mutating their arithmetic operators $A = \{+ ,  -, *, /\}$
	\STATE  $\{\varepsilon^{'}_t \} \leftarrow$ \textit{MO-CMA-ES ($\varepsilon$, A)} 
	\STATE $P^{'}_{k} \leftarrow$ \textit{Background-Subtraction-Method($\varepsilon^{'}_{k}, X)$}
	\STATE \textit{lbp-scores}$\leftarrow$ \{$P^{'}_{k}$, $\varepsilon^{'}_{k} \} $
	\ENDFOR	
	\STATE // Select the best equation
	\STATE  \{$P^{'}$, $\varepsilon^{'} \} $ = $\arg\max(\textit{lbp-scores})$
	\STATE \textbf{Output:}  Best $\varepsilon^{'}$ equation and its accuracy
	\end{algorithmic}
\end{algorithm}

\subsection{Search Strategy by MO-CMA-ES to Discover the Best Equation}
\label{sec:search_strategy}
With a search space designed, $E = \{\varepsilon_1,  \varepsilon_2, \dots,  \varepsilon_K \}$, where $\varepsilon$ expresses each equation structures and $K$ is a user parameter that determines the number of elements $\varepsilon \subseteq E$, the MO-CMA-ES looks for the best LBP equation by mutating the arithmetic operators $(+ ,  -, *, /)$ of each equation $\varepsilon \subseteq E$ resulting in a new set of mutated equations given by $E^{'} = \{\varepsilon^{'}_1, \varepsilon^{'}_2, \dots,  \varepsilon^{'}_K \}$. We estimate the performance of each LBP equation by using a background subtraction algorithm called Texture BGS~\citep{heikkila:2006}. Alows us to distinguish moving objects from the background of a set of videos. Finally, the  $\varepsilon^{'} \subseteq E^{'}$ that presented the maximum accuracy $\{ \arg\max_{j \in 1, \ldots, K} P^{'}_{j}(x)\}$ is chosen to deal with a particular challenge that is commonly encountered in complex scenes (e.g., changes in lighting, dynamic backgrounds, camera jitter, noise, shadows and others). The $P^{'}$ represents a set of accuracy for each $\varepsilon^{'} \subseteq E^{'}$. A complete description of all steps of the proposed framework is presented in Algorithm \ref{alg:structure_discovery}. The Figure 3 shows a brief illustration of this step of our framework. An introduction of the MO-CMA-ES is  described below. A state-of-the-art with a detailed description of the MO-CMA-ES can be found in~\citep{igel:2007, hansen:2011}.

\subsubsection{Multi-Objective Covariance Matrix Adaptation Evolution Strategy (MO-CMA-ES)}
\label{subsec:search_strategy_best_equation}
 Multi-Objective Covariance Matrix Adaptation Evolution Strategy (MO-CMA-ES) is a search strategy algorithm that ranks the elements in a population $A$ of candidate solutions according to their level of non-dominance. It is based on two hierarchical criteria: \textit{Pareto ranking and hypervolume contribution}. Let the non-dominated solution in $A$ be represented by ndom(A) = $\{a \in \mathrm{A} \mid \nexists a^{'} \in \mathrm{A}: a^{'}  \prec a \}$, where  $\prec$ indicates the Pareto-dominance relation. The Pareto front of $A$ is then given by  $\{{(f_{1}(x), \ldots, f_{m}(a)) \mid a \in {\text{ndom(A)}}}\}$, where the $f_{i}$ are the $m$ real-valued objective functions. 
 Initially, the Pareto by interactively process ranks the elements in ndom(A)  a level of non-dominance getting rank 1. The other ranks of non-dominance continue until all points of $A$ have received a Pareto rank. Next, the solutions  that have  the same level of non-dominance are ranked by considering the \textit{contributing hypervolume}. The hypervolume measure or $\mathcal{S}$-metric \citep{zitzler:1998} is determined in as follows:

\begin{equation}
\label{eq:4}
\mathcal{S}_{f^{ref}}(A) =  \land \left( \bigcup_{a \in A} \left[ f_1 (a), f_{1}^{ref} \right] \times \cdot \cdot \cdot \times  \left[ f_m (a), f_{m}^{ref} \right]  \right)
\end{equation}

where $f^{ref} \in \mathbb{R}^{m}$ indicates an appropriately selected reference point and $\land (\cdot)$ (i.e., the volume) being Lebesgue measure. The contributing  hypervolume of a point $a \in A^{'}$= ndom(A) is given according by $\bigtriangleup_{\mathcal{S}}(a, A{^{'}}) =  \mathcal{S}_{f^{ref}}(A^{'}) - \mathcal{S}_{f^{ref}}(A^{'} \backslash \{a\})$. Element $a$ with the lowest contributing hypervolume is assigned to contribution rank 1. 

Finally, in the MO-CMA-ES version considered in this paper, an individual offspring $ a^{'(g + 1)}_{i}$ is determined to be successful if selected for the next parent population $P^{(g+1)} = \{ Q^{g}_{\prec : i} \mid 1 \leq i \leq \mu \}$:

\begin{equation} 
\label{eq:9}
succ_{Q^{(g)}}^{P} \left( a_{i}^{(g)}, a^{'(g+1)}_{i} \right) = 
  \begin{cases}
    1 & \text{if $a_{i}^{'(g + 1)} \in Q^{(g+1)}$} \\
    0 & \text{otherwise}
  \end{cases}
\end{equation}

where succ$_{Q^{(g)}}^{P} \left( a_{i}^{(g)}, a^{'(g+1)}_{i} \right)$ in  Eq.(\ref{eq:9})  is success indicator. It evaluates as one if the mutation that created  $ a^{'(g + 1)}_{i}$  is considered successful and zero otherwise.

\section{Experimental Results}
\label{sec:experimental_reults}

In this section, we evaluate each component of our proposed method by conducting different experiments. First, we start by finding the set of optimal hyperparameters to train a VAE instance that generates a set of valid equations that are different from those within the training set. Therefore, we defined an evaluation measure called number of \emph{Unseen} \& \emph{Valid Equations} (UVE), and our main objective in this first experiment was to find a configuration for the VAE instance that maximizes the UVE measure. Each VAE instance with specific hyperparameters generates a set of equations, so its validity is verified by a regular expression. As a search strategy algorithm, we choose the MO-CMA-ES due to its simplicity and scalability to higher dimensional problems. In addition, it is parallelizable and easily distributed. This can be a very important factor in AutoML approaches that spend a lot of computational time. 
To perform the experiments optimally, we use the \emph{Proactive Machine Learning (PML)} \footnote{https://www.activeeon.com/products/proactive-machine-learning}. It is an open source solution developed by \href{https://www.activeeon.com}{Activeeon} that allows researchers to easily automate and orchestrate AI-based workflows, scaling-up with parallel and distributed execution in any type of infrastructure: on premise, hybrid, multi-cloud or HPC (High Performance Computing), by taking into account full computing capacity, from CPU to GPU/TPU/FPGA. PML allows hyperparameter optimization on a large scale, and users are free to implement their own machine learning algorithms using different programming languages such as Python, MATLAB, R, C, Java etc.

We implemented the proposed framework using PyTorch 1.0 and Python 3.6. The experiments were conducted on an NVIDIA GeForce RTX 2070 graphics card with 3.5 GHz Intel i7-3770K CPU that consists of 4 cores and 8 threads, and 16GB of RAM. First, we ran 150 interactions of two parallel instances of the VAE with different hyperparameters using the PML platform. VAE models are based on the Gated Recurrent Unit (GRU)~\citep{cho:2014}. It's an enhanced version of the standard recurrent neural network that aims to solve the vanishing gradient problem. The GRU can also be known as a variation on the Long short-term memory (LSTM)~\citep{hochreiter:1997} because both are  similarly designed and, in some cases, produce equal results. Nevertheless, we choose to use GRU instead of LSTM because it has less training parameters and therefore uses less memory, running time and trains faster. We trained our model into 150 epochs using an early stopping mechanism. Therefore, we seek to minimize the KL-divergence and the reconstruction loss. In Table \ref{table:1}, we tabulate the hyperparameters used in the VAE network and in the training stage. The first column contains the definitions of each parameter. In the second column the range value that each of these parameters can assume and finally the last column shows the combination of the best values for each parameter. The set of the best parameter values were used the configuration of our VAE that will be responsible for generating our search space containing a set of equations. Note that the \emph{choice} represents a list of possible values or parameters for each variable.

\begin{table}[t]
\centering
\scalebox{0.72}{
\begin{tabular}{| l | r  | r |} 
 \hline
 \textbf{Hyperparameter} & \textbf{Range Values} & \textbf{Best Values} \\[0.5ex] 
 \hline\hline
 \toprule
    \multicolumn{3}{c}{\textbf{VAE network parameters}}\\
     \hline
ENC\_HIDDEN - Number of features in the hidden of the encoder  & \emph{choice}([125, 256, 512]) & 125  \\ 
DEC\_HIDDEN - Number of features in the hidden of the decoder  & \emph{choice}([512, 800])  & 512  \\ 
ENC\_LAYERS  - Number of recurrent layers of the encoder  & \emph{choice}([1, 2, 4, 6])  & 6   \\
DEC\_LAYERS - Number of recurrent layers of the decoder  & \emph{choice}([1, 2, 4, 6])  & 1   \\
ENC\_DROPOUT - Dropout rate  of the encoder  &  \emph{choice}([0.01, 0.02, 0.01, 0.1,  0.2]) & 0.1 \\
DEC\_DROPOUT - Dropout rate of the decoder  &  \emph{choice}([0.01, 0.02, 0.01, 0.1,  0.2])  & 0.01   \\
\toprule
    \multicolumn{3}{c}{\textbf{Training parameters}}\\
     \hline

 N\_BATCH  -  Samples per batch to load & \emph{choice}([32, 64, 512])  & 32 \\  
 LEARNING\_RATE - Learning rate & \emph{choice}([0.001, 0.005])  & 0.005 \\  
 OPTIMIZER - Optimization algorithms &  \emph{choice}([Adam, Adadelta, RMSprop])   & RMSprop   \\  [1ex] 
 \hline
\end{tabular}}
  \begin{tablenotes}
     \tiny
      \item   \textbf{${}^\star$ If (\textit{ENC\textunderscore LAYERS}) or (\textit{DEC\textunderscore LAYERS}) $>$ 1, becomes a bidirectional GRU otherwise unidirectional GRU.}
    \end{tablenotes}
\caption{\small List of hyperparameters used in the VAE network and in the training stage. In the second column the range value that each of these parameters can assume and finally the last column shows the combination of the best values for each parameter. The set of the best parameter values is used as the configuration of our VAE that will be responsible for generating our search space containing a set of equations.}
\label{table:1}
\end{table}

\begin{sidewaysfigure}[t]
\label{fig:graphvae}
 \includegraphics[width=\linewidth]{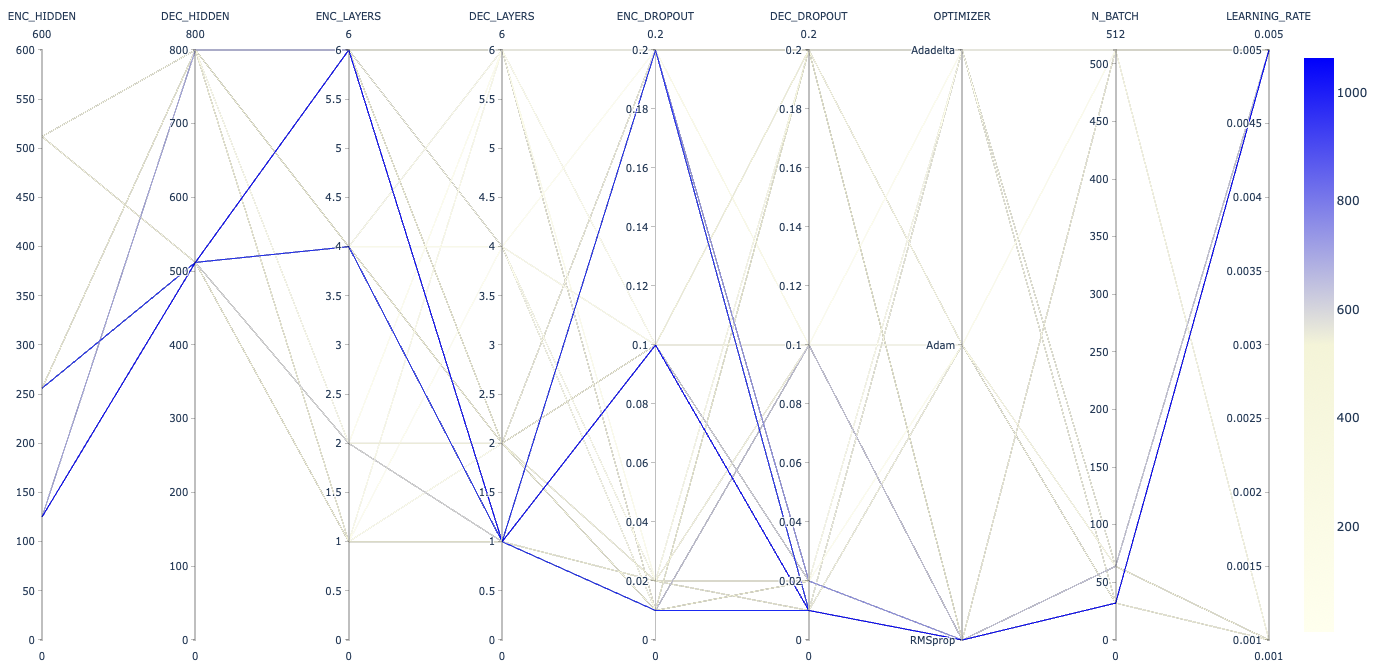}
 \caption{\small Different hyperparameters used in the VAE. An overview of how the different instances of VAE behave when varying the values of their hyperparameters to maximize the UVE measure.}
\end{sidewaysfigure}

In Figure \ref{fig:graphvae}, we can see an overview of how the different instances of VAE behave when varying the values of their hyperparameters to maximize the UVE measure. Some hyperparameters values are important, while others could even be eliminated, thereby reducing our initial set of hyperparameters. For instance, the \textit{ENC\_HIDDEN:} {125, 256}, \textit{DEC\_HIDDEN:} {512}, \textit{ENC\_LAYER:} {4, 6}, \textit{DEC\_LAYER:} {1}, \textit{ENC\_DROPOUT:} {0.01, 0.1, 0.2}, \textit{DEC\_DROPOUT:} {0.01}, \textit{OPTIMIZER:} {RMSpop}, \textit{N\_BATCH:} {32} and \textit{LEARNING\_RATE:} {0.05} were parameter values that contributed to maximize the number of UVE. Figure 5 
shows the importance of each hyperparameter when trying to maximize the UVE measure. As we can see the \textit{ENC\_DROPOUT, ENC\_LAYERS and DEC\_DROPOUT} were the hyperparameters that most impacted on VAE training. This can be explained by the fact that the dropout layer is very important in Generative Network architectures because they pre-
\newpage

\clearpage

vent all neurons in a layer from synchronously optimizing their weights preventing all the neurons from converging to the same goal, thus decorrelating the weights. Another hyperparameter that normally positively impacts the performance of the generative neural models is the number of recurrent layers in the encoder. A study presented by \cite{wu:2016} showed that the 2-layer bidirectional encoder performed slightly better than other tested configurations. However, in Figure \ref{fig:graphvae} we can see that 4 or 6 bidirectional layers provided better results to solve our problem, thus reinforcing the theory that deeper networks achieve better performance than shallow networks. A curious fact is that the number of decoder layers did not have a major positive impact on the performance of the VAE models in this work. This behavior could be explained by the fact that the encoder layers as well as its dropouts played a more important role during the equations generation than the decoder layers. Once we define a set of optimal hyperparameters for VAE model, it generated a set of 305 valid and unique LBP equations structures. They are based on the LBP modified by \cite{heikkila:2006} for background modeling. The authors proposed to modify the threshold scheme of the original LBP equation, replacing the term $s(Z-C)$  with the $s(Z-C+a)$ where $C$ corresponds to the gray value of the center pixel of a local neighborhood, $X$ to the gray values of $P$ pixels equally spaced on a circle of radius R and $a$ is a small value. Note that in our experiments $a$ is defined as 0.01. The modified LBP improves the original LBP equation in image areas where the gray values of the neighboring pixels are very close to the center pixel, e.g. sky, grass, etc. The MO-CMA-ES changed each LBP equation using $4^{n}$ permutations, where  $4$ is the number of the types of arithmetic operations $[+ ,  -, *, /]$) and $n$ is number of arithmetic operators presented in the equation. However, to prevent the generation of exponential mutations, we limited our experiment to $4^{5}= 1024$ mutations. In order to verify if the mutated equations are useful for dealing with different challenges found in real-world scenarios, we consider as our performance estimation strategy a background subtraction algorithm called Texture BGS proposed by~\cite{heikkila:2006} and a set of real-world scene sequences from CDnet-2014 data set~\citep{wang:2014}. A brief description of the Texture BGS algorithm and CDnet-2014 are presented below:

\begin{itemize}
\item \textbf{Texture BGS method}: it is a non-parametric approach that models each pixel as a group of adaptive LBP histograms that are calculated over a circular region around the pixel. Initially, the LBP histogram of the given pixel is calculated from the new video frame and compared to the current model histograms using an intersection measure of the histogram. Then, the histogram's persistence in the model is used to decide whether the histogram models the background or not. The higher the weight, the more likely it is a background histogram. At the end of the updating procedure, the model's histograms are ordered in descending order according to their weights, and the first histograms are select as background histograms. 

\item \textbf{CDnet-2014 data set}: it is the largest data set with videos that contains several motion and change detection challenges, in addition to typical indoor and outdoor scenes that are found in most surveillance, smart environments, and video analytics applications. The CDnet data set contains 53 annotated videos from
various real-world scenarios. These videos are categorized into 11 categories such as \emph{bad weather, low framerate, night videos, PTZ, thermal, shadow, intermittent object motion, camera jitter, dynamic background, baseline and turbulence}. 
\end{itemize}

\begin{figure}[t]
\centering
\label{fig:hyperimportance}
 \includegraphics[width=0.75\textwidth]{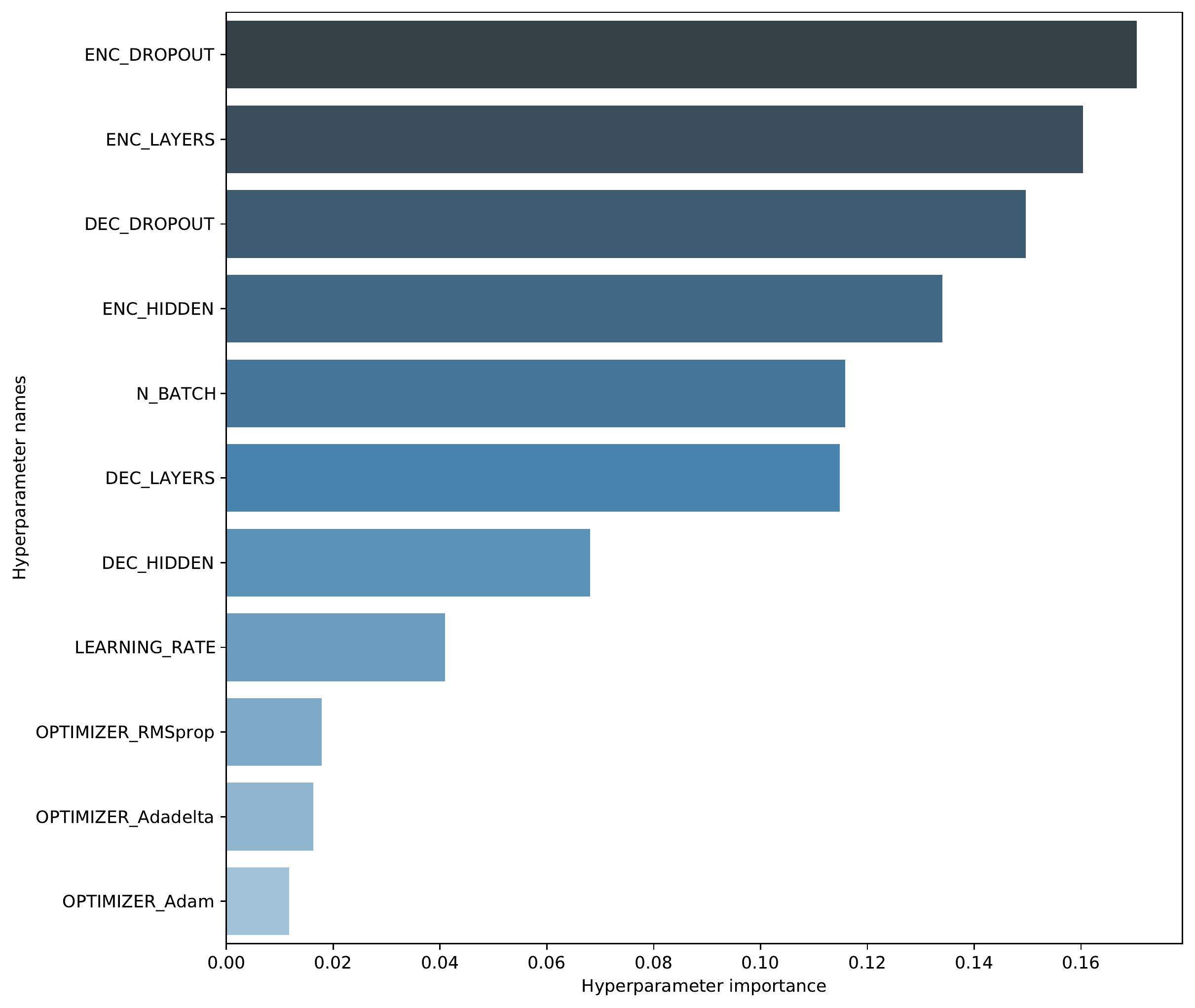}
 \caption{\small Hyperparameters Importance. The \textit{ENC\_DROPOUT, ENC\_LAYERS} and \textit{DEC\_DROPOUT} were the hyperparameters that had the greatest impact on training.}
\end{figure}

Each mutated LBP equation was evaluated using the Texture BGS method in \emph{'peopleInShade', 'snowFall', 'canoe', 'busStation', 'skating'}, and \emph{'fall'} sequences with resolutions 180 x 120 pixels. The description of each scene and its main challenges are presented in Table \ref{tab:challenges}. We limited the number of sequence for each scene around 7 $\sim$ 40 frames due to the limitation of computational resources, since our approach has high computational complexity and computation time. All equations were performed in parallel thanks to the PML platform. We present the visual results on individual frames from six different  scenes: \emph{'peopleInShade'} (frame \#318), \emph{'snowFall'} (frame \#2758), \emph{'canoe'} (frame \#904), \emph{'busStation'} (frame \#300),  \emph{'skating'} (frame \#1845) and \emph{'fall'} (frame\#3987) of the  CDnet-2014 data set. 
Figures 6 and 7 show the foreground detection results using the Texture BGS and the proposed method, respectively. They were shown without any  post-processing technique. 
The results obtained by using the best LBP equations discovered by the proposed method (see Table \ref{tab:challenges}) clearly appears to be less sensitive to background subtraction challenges and are able to  detect moving objects with fewer false detection, especially in the \emph{'canoe'} video which is a complex scene with a strong background motion and in the \emph{'skating'} scenes that presents challenges such as low-visibility winter storm conditions and snow accumulation.  Next, given the ground truth data, the accuracy of the foreground segmentation is measured  using  three  classical  measures:  recall,  precision and  F-score. Table 3	
shows the  proposed framework evaluated on in the six scenes.  The  best scores are  in  bold.  The  proposed  approach presented  the  best  scores  for \emph{'peopleInShade', 'snowFall', 'canoe', 'busStation'} and \emph{'skating'} while it performing the worst score for the \emph{'fall'} scene. However, we can improve the  results of our proposed method by conducting an exhaustive search to find the best LBP equation for this scene. The authors of the Texture BGS method reported in their work that shadow turned out to be an extremely difficult problem to solve for this method. However, as we can see in Table 3, the best LBP equations discovered by our approach to dealing with the shadow problem in \emph{'peopleInShade'} and \emph{'busStation'} scenes were able to improve the results considerably. 

We also evaluate the best LBP equations discovered by our algorithm in unseen sequences. Therefore, we selected around 100 $\sim$ 150 sequences for each scene. The results are shown in Figure 8. From the chart, we can see that although the F1 score has decreased for most scenes in unseen sequences (orange bar), the best LBP equations are still adequate to deal with unseen sequences over time.  In contrast, the F1 score dropped drastically from 0.8825 \%, to 0.3809 \% for the \emph{'busStation'} scene. The best LBP equation discovered by our algorithm does not seem to be adequate for \emph{'busStation'} scene over time. To help us understanding the performance of the \emph{'busStation'} scene, we performed a new experiment adding 27 more new sequences during the mutation process and our algorithm discovered a new best equation: $(Z/C)/(Z/C)/(Z*(Z/C)-(Z+C))-a$.
When applying this better equation to unseen sequences, it seems to deal better with the challenges presented in the \emph{'busStation'} scene by increasing the F1 score from 0.3809\% to 0.4483\%.

In  general, we  can  see  that  the proposed framework is capable of finding useful LBP equations to deal with different challenges encountered in a given scene. Although one scene has similar challenges to the other, our experiments showed that the best equation for dealing with them will not be the same, since each one is unique and the dynamics of the scene also contributes to the choice of the best equation. This can be noticed by the couples of scenes [\emph{'peopleInShade'} and \emph{'busStation'}], [\emph{'canoe'} and \emph{'fall']} and [\emph{'snowFall'} and \emph{'skating'}] that have similar challenges and yet the best equation found out for each of them by our approach is different (see Table 2). Furthermore, when performing experiments in unseen sequences, we can see that the best LBP equation that seemed to be adequate to deal with the  \emph{'busStation'} scene initially does not seem to be appropriate over time. By increasing the number of sequences during the mutation process, a new better equation was discovered indicating that the number of sequences used in the mutation process can impact the performance of our framework.

Finding a specific equation for a given scene manually is a laborious task that requires a in-depth knowledge of the scene and a trial-error process by experts. The approach presented in this paper can be very useful saving human research time. Finally, the best mathematical LBP equations found out in this work may be different for the same scene, especially if a different background subtraction method is used or our approach is applied to another type of domain. In addition, the results presented in this paper are an non-exhaustive search for the best LBP, due to limited computational resources. Therefore, the approach presented in this paper is especially ideal for discovering the best mathematical equations for solving specific problem in computer vision field.

\begin{figure}[t]
\label{fig:visual_result}
 \includegraphics[width=0.95\textwidth]{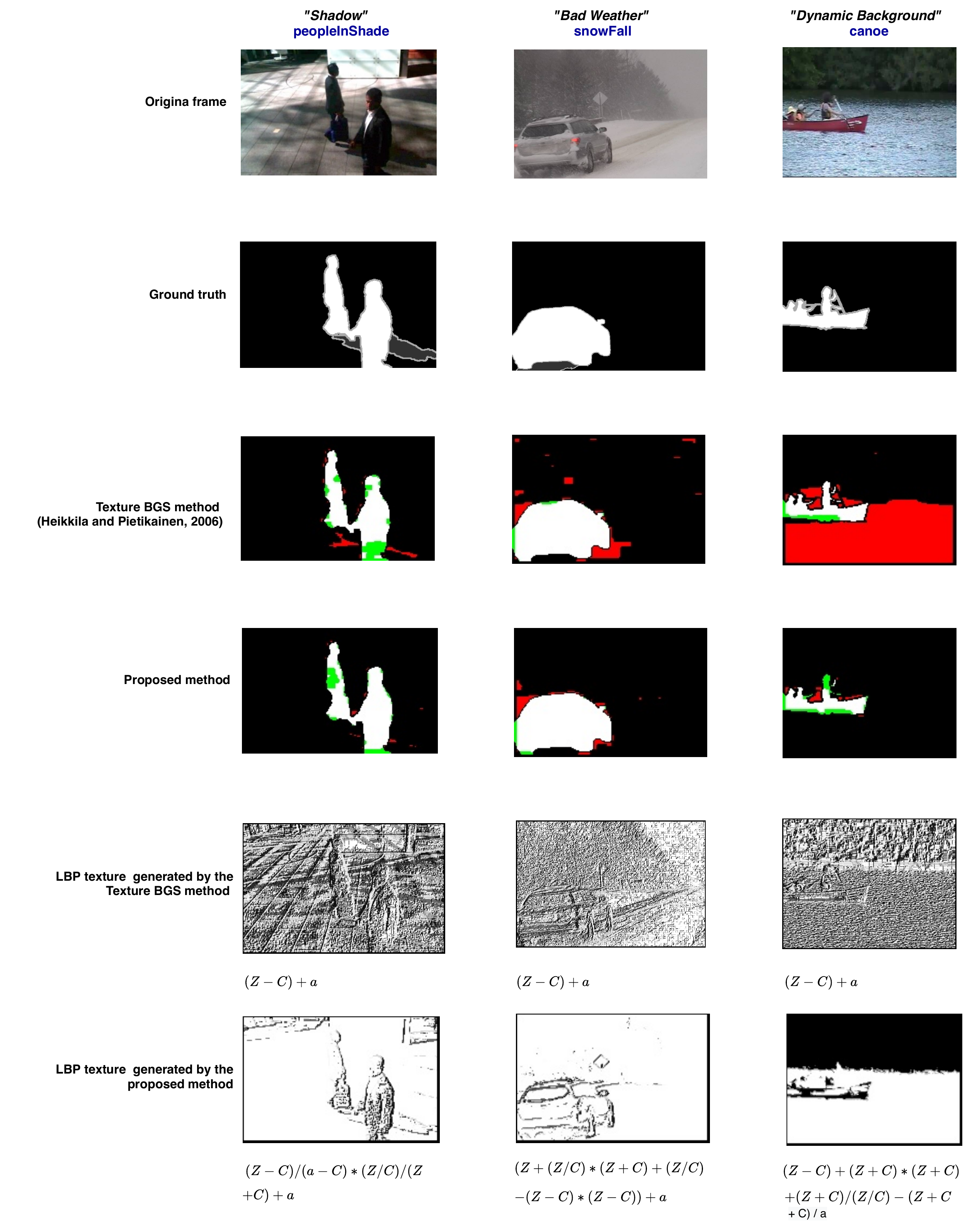}
 \caption{\small Background subtraction results using the CDnet-2014 data set. From top to bottom: (a) Original frame, (b) Ground truth, (c)  Texture BGS method, (d) Proposed method, and (e) LBP texture. The true positives (TP) pixels are in white, the true negatives (TN) pixels are in black, the false positives (FP) pixels are in red, and the false negatives (FN) pixels are in green.}
\centering
\end{figure}

\begin{figure}[t]
\label{fig:visual_result2}
 \includegraphics[width=0.95\textwidth]{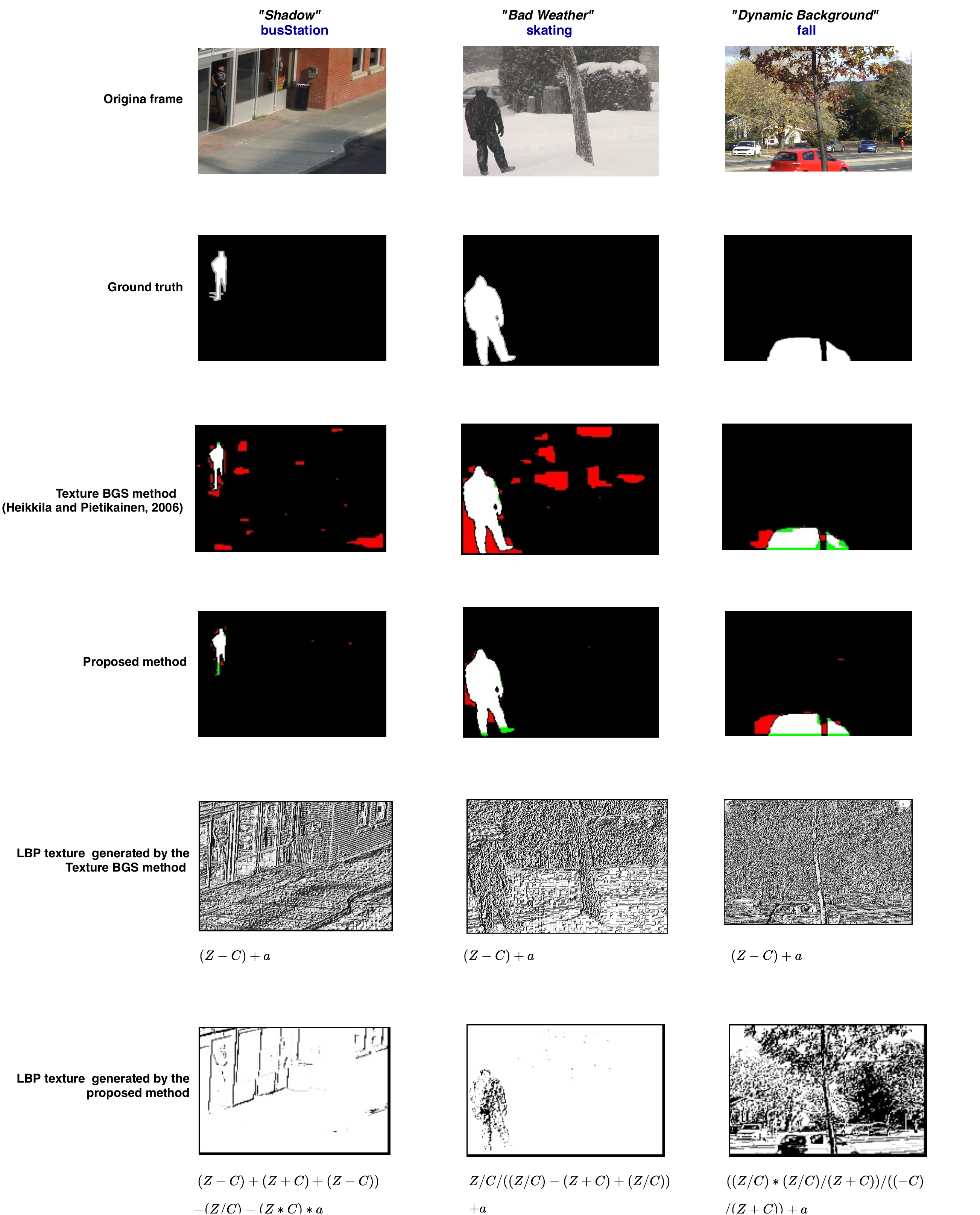}
 \caption{\small Background subtraction results using the CDnet-2014 data set. From top to bottom: (a) Original frame, (b) Ground truth, (c)  Texture BGS method, (d) Proposed method, and (e) LBP texture. The true positives (TP) pixels are in white, the true negatives (TN) pixels are in black, the false positives (FP) pixels are in red, and the false negatives (FN) pixels are in green.}
\centering
\end{figure}

\clearpage

\begin{table}[t]
\centering
\scalebox{0.92}{
\begin{tabular}{ |p{2.6cm}|p{4.5cm}|p{3.8cm}|p{3.8cm}|  }
 \hline
 \textbf{Videos}     & \textbf{Challenge} & \textbf{LBP structure} & \textbf{Best LBP equation}\\
 \hline
  \hline
\multirow{6}{6em}{\centering\em peopleInShade}  & Consists of pedestrian walking outdoor. The main challenges are: hard shadow cast on the ground by the walking persons and illumination changes.  & (Z o C) o (a o C) o (Z o C) o (Z o C) o a  &   (Z - C) / (a - C) * (Z / C) / (Z + C) + a \\

 \hline
\multirow{6}{6em}{\centering\em snowFall}  & Contains a traffic scene in a blizzard. The main challenges are: low-visibility winter storm conditions, snow accumulation, the dark tire tracks left in the snow. & (Z o (Z o C) o (Z o C) o (Z o C) o (Z o C) o (Z o C)) o a &    (Z +( Z / C) * (Z + C) + (Z / C) - (Z - C) * (Z - C)) + a \\

 \hline
\multirow{6}{6em}{\centering\em canoe}  & Shows people in a boat with strong background motion. The main challenges are: outdoor scenes with strong background motion.  & (Z o C) o (Z o C) o (Z o C) o (Z o C) o  (Z o C) o (Z o C o C) o a &   (Z - C) + (Z + C) * (Z + C) + (Z + C) / (Z / C) - (Z + C + C) / a \\
 \hline
 \multirow{6}{6em}{\centering\em busStation}  & Presents people waiting in a bus station. The main challenges are: hard shadow cast on the ground by the walking persons. & (Z o C) o (Z o C) o (Z o C) o (Z o C) o (Z o C) o a &    (Z - C) + (Z + C) + (Z - C) - (Z / C) - (Z * C) * a \\
 \hline
\multirow{6}{6em}{\centering\em skating}  & Shows people skating in the snow. The main challenges are: low-visibility winter storm conditions and snow accumulation. & Z o C o ((Z o C) o (Z o C) o (Z o C)) o a &    Z / C/ ((Z / C) - (Z + C) +(Z / C)) + a\\
 \hline
\multirow{6}{6em}{\centering\em fall}  & Show cars passing next to a fountain. The main challenges are: outdoor scenes with strong background motion.  & ((Z o C) o (Z o C) o (Z o C)) o ((o C) o (Z o C)) o a  &    ((Z / C) * (Z / C) / (Z + C)) / ((-C) / (Z + C)) +  a \\
 \hline
\end{tabular}}
\caption{\small List of the best equations for dealing with different challenges encountered in real-world scenarios. From the left to the right:  (a) different types of scenes, (b) with their main challenge descriptions, (c) the best LBP structure, and (d) the best LBP equation.}
\label{tab:challenges} 
\end{table}


   \newcommand{\clarg}{25mm}
   
	\begin{table}[t]
	\centering
	\label{tab:performance} 
	\begin{tabular}{|l|l|c|c|c|}
	\hline
	{\bf Videos}	&  {\bf Method}&	{\bf Precision}	& {\bf Recall}	& {\bf F-score} \\
	\hline
	\hline
	\multirow{2}{\clarg}{\centering\em peopleInShade}		
	&  Texture BGS   &   0.7339 &  {\bf0.9009}  &  	0.8088 \\
	& Proposed approach           & {\bf0.8211} & 0.8988 & {\bf0.8582} \\
	\hline
	\multirow{2}{\clarg}{\centering\em  snowFall}
	&  Texture BGS   &   0.5970 &  {\bf0.9384}  &  	0.7298 \\
	& Proposed approach           & {\bf0.8673} & 0.91638 & {\bf0.8911} \\
	\hline
	\multirow{2}{\clarg}{\centering\em  canoe}
	&  Texture BGS      &   0.1106 &  \textbf{0.8183}  &  0.1949 \\
	& Proposed approach &   \textbf{0.8710} &  0.5535  &  \textbf{0.6769} \\	
	\hline
	\multirow{2}{\clarg}{\centering\em  busStation}
	&  Texture BGS   &   0.2747 &  {\bf0.9695}  &  	0.4282 \\
	& Proposed approach           & {\bf0.8792} & 0.8859 & {\bf0.8825} \\
	\hline
	\multirow{2}{\clarg}{\centering\em  skating}	
	&  Texture BGS   &   0.2666 &  {\bf0.9161}  &  	0.4130 \\
	& Proposed approach           & {\bf0.9213} & 0.8520 & {\bf0.8853} \\
	\hline
	\multirow{2}{\clarg}{\centering\em  fall}	
	&  Texture BGS   &   {\bf0.5453} &  0.7904  &  	{\bf0.6453} \\
	& Proposed approach           & 0.4625 & {\bf0.8190} & 0.5912 \\
	\hline
	\end{tabular}
	    \caption{\small Performance using the CDnet-2014 data set.}
	\normalsize
	\vspace{-5mm}
	\end{table}

\begin{figure}[H]
\centering
\label{fig:validation}
 \includegraphics[width=0.86\textwidth]{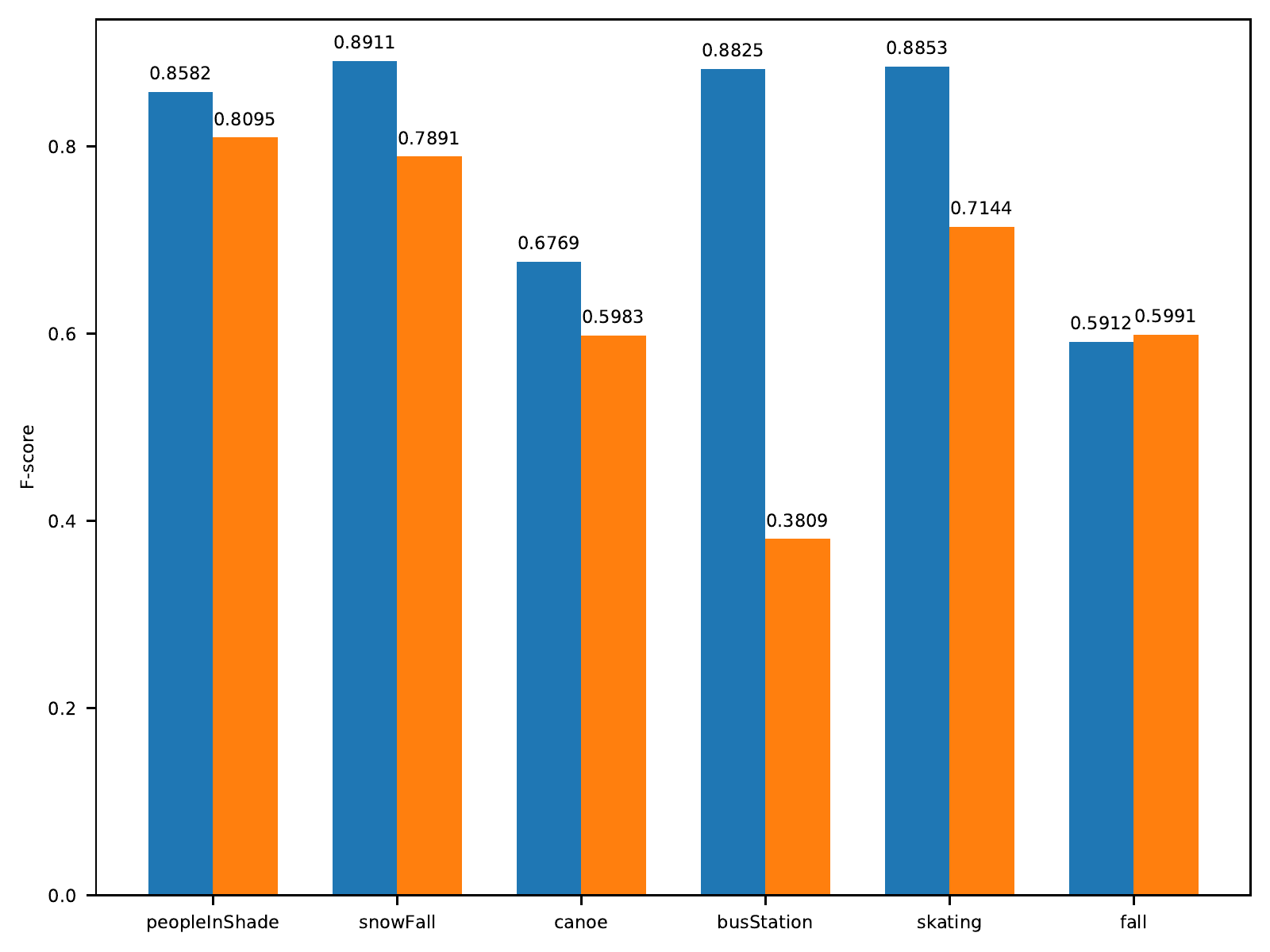}
 \caption{The F1 score obtained by our proposed approach in scenes from \emph{'peopleInShade', 'snowFall', 'canoe', 'busStation', 'skating'}, \emph{fall}. The blue bar on the chart represents the results presented by the best equations discovered during the mutation process, while the orange bar represents the result obtained by the best equations discovered in unseen scenes. Observe that while the F1 score has decreased for most scenes in unseen sequences (orange bar), the best LBP equations are still adequate to deal with unseen sequences over time.  In contrast, the F1 score dropped dramatically from 0.8825 \%, to 0.3809 \% for the \emph{'busStation'} scene.}
\end{figure}

\newpage

\section{Conclusion}
\label{sec:conclusion}

In this paper, we present a novel framework for discovering the appropriate Local Binary Patterns to deal with the main challenges encountered in real-world scenarios. The proposed framework can reduce human bias and time by proposing a search space designed by a generative network instead of hand-designed features. In addition, it can discover equations that we may never have thought of yet. Experimental results in video sequences show the potential of the proposed approach and its effectiveness in discovering the best equation for dealing with the problems found in a particular complex scene. Future work includes formulating new structures of equations other than LBP and carrying on experiments on scenes with different challenges. Moreover, the method developed in this paper could be extended to neural architecture search. This could be done by generating and combining basic mathematical structures as building blocks, enabling the construction of a variety of neural architectures.

\bibliography{paper}

\end{document}